\documentclass[letterpaper, 10 pt, journal, twoside]{ieeetran}
\IEEEoverridecommandlockouts
\usepackage{cite}
\usepackage{amsmath,amssymb,amsfonts}
\usepackage{algorithmic}
\usepackage{multirow}
\usepackage{graphicx}
\graphicspath{{./pics/}}%
\usepackage{textcomp}
\usepackage{xcolor}
\usepackage{gensymb}
\usepackage{threeparttable}
\usepackage{ulem}
\def\BibTeX{{\rm B\kern-.05em{\sc i\kern-.025em b}\kern-.08em
    T\kern-.1667em\lower.7ex\hbox{E}\kern-.125emX}}
\begin{document}


\title{Design and Verification of a Novel Triphibian Robot
\thanks{$\ddagger$ These two authors contribute equally.}
\thanks{$^1$Kaiwen Xue, Chongfeng Liu and Huihuan Qian are with the Shenzhen Institute of Artificial Intelligence and Robotics for Society (AIRS), The Chinese University of Hong Kong, Shenzhen, Guangdong, China}
\thanks{$^2$Shiqi Yang, Kaiwen Xue, Minen Lv, Yingtai Xu, Yiying Lu, Chongfeng Liu and Huihuan Qian are also with the School of Science and Engineering, The Chinese University of Hong Kong, Shenzhen, Guangdong, China.}
\thanks{$^3$Jingyi Yang is with the School of Data Science, The Chinese University of Hong Kong, Shenzhen, Guangdong, China.}
\thanks{$\dagger$The corresponding author is Huihuan Qian.}
}

\author{Shiqi Yang$^{2}\ddagger$, Kaiwen Xue$^{1,2}\ddagger$, Minen Lv$^{2}$, Yingtai Xu$^{2}$, Jingyi Yang$^{3}$, Yiying Lu$^{2}$, Chongfeng Liu$^{1,2}$ and Huihuan Qian$^{1,2\dagger}$ 
}

\maketitle

\begin{abstract}
Multi-modal robots expand their operations from one working medium to another, land to air for example. The majorities of multi-modal robots mainly refer to platforms that operate in two different media. However, for all-terrain tasks, there are seldom research to date in the literature. Generally, locomotions in different working media, i.e. land, water and air, require different propelling actuators, and thus the triphibian system becomes bulky. To overcome this challenge, we proposed a triphibian robot and provide the robot with driving forces to perform all-terrain operations in an efficient way. A morphable mechanism is designed to enable the transition between different motion modes, and specifically a cylindrical body is implemented as the rolling mechanism in land mode. Detailed design principles of different mechanisms and the transition between various locomotion modes are analyzed. Finally, a triphibian robot prototype is fabricated and tested in various working media with both mono-modal and multi-modal functionalities. Experiments have verified our platform, and the results show promising adaptions in future exploration tasks in various working scenarios.
\end{abstract}


\section{Introduction}
Robotic platforms aid human beings in performing various tasks in different working environments, such as ground vehicles transporting on land, unmanned surface vehicles patrolling on water surface and aerial vehicles surveying in the air. Typically, the majorities of robots are mono-modal and focus on their specific tasks within one type of media (i.e. land, water or air) \cite{review-multi-modal}. However, for the robots working in the complex field, the challenges come from divergent environments in different mediums, such as various land terrains, water surfaces and even obstacles to jump or go over, which requires more flexible robot designs \cite{aerial-aquatic-morph-design}\cite{aerial-aquatic-morph-tmech}. Therefore, multimodal robots with more than one locomotion method will enable themselves with greater adaptation to overcome the challenges in various working scenarios with different mediums, such as \textcolor{black}{full perception} with water and air media transition \cite{aerial-underwater-HUAUV}, and exploration tasks traversing or flying over high obstacles \cite{land-air-robot1}-\cite{land-air-robot3}. Thus, more recent research efforts have been put into the multimodal robotic platforms development and the working environment properties expansion.
\begin{figure}[htbp]
	\centering
	\includegraphics[{width=\linewidth, height = 0.65\linewidth}]{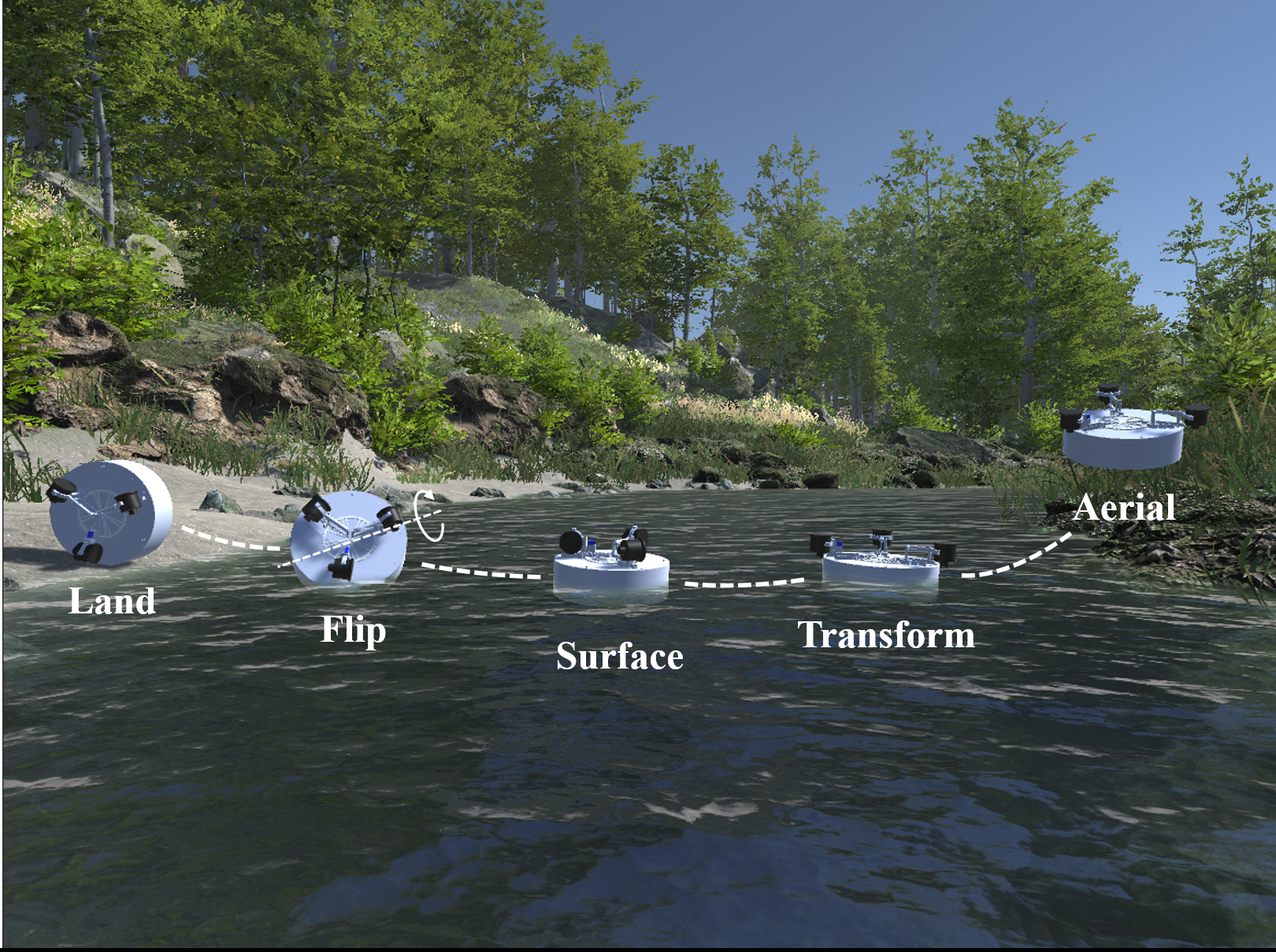}
	\caption{Concept view of the proposed robot. The robot starts with the land mode and rolls down to the water surface with the surface mode. Further, it transforms to the aerial mode to leave the water environment.}
	\label{fig-intro}
\end{figure}
\par For the different locomotion modes, they can be decomposed broadly into three types: terrestrial (or land), aquatic \textcolor{black}{(or surface)} modes \cite{review-multi-modal}, and aerial, and the multi-modal functionalities can be typically classified by the transition between different locomotion modes, i.e. aerial-aquatic locomotion, land-aerial, land-aquatic locomotion, and even triphibian modes with all three locomotion modes.
For aerial-aquatic locomotion, the difficulty mainly comes from the fact that fluid density in air and water poses great variation and thus affects the locomotion mechanism design. Currently, to fill this gap, there are mainly two kinds of wing-type designs (flapping wing-type designs or passive fixed-wing-type designs) and two kinds of rotor-type designs (with multirotor devices or morphable devices). The first kind of wing-type design is the flapping wing-type design. Some robots based on such design \cite{aerial-aquatic-flap-insect}\cite{aerial-aquatic-flap-science} adopted versatile flapping propulsive strategies for the water entry and exit operation. Some wing-type multi-modal robots utilize air to enhance their flying \cite{aerial-aquatic-fixed-wing-saill}. The second wing-type design is the passive fixed-wing-type design. The robots based on such design \cite{aerial-aquatic-fixed-wing-buoyance} adjust their buoyance by passively flooding or draining their wings. Different from wing-type design robots, the rotor-type design robots mainly take advantage of the multirotor platforms for operation ease \cite{aerial-underwater-HUAUV-rotor}, and control strategies from the multirotor unmanned aerial vehicles \cite{aerial-underwater-HUAUV}\cite{aerial-aquatic-JFR}. Furthermore, rotor-type designs with morphable devices can further achieve an optimized aquatic operation by morphing its rotors/thrusters direction \cite{aerial-aquatic-morph-design}\cite{aerial-aquatic-morph-tmech}. 
For the land-aerial robotic platforms, typically they employ wheels (on land) and rotor-like devices (in the air) together with some transformation devices \cite{land-air-robot1}-\cite{land-air-flystar}. 
For the land-aquatic designs, \cite{land-aquatic-AmphiBotI} proposed a snake-like robot by mimicking the gaits of a snake both on land and in water. \cite{land-aquatic-wheel} designed a wheel-propeller-fin mechanism for land and aquatic operations.

\textcolor{black}{For the triphibian platform designs, \cite{tri-robot-1} and \cite{tri-robot-2} both mixed the wheels, quadrotors, and a buoyance mechanism (hollow body or vacuum tire) as a platform for all three different locomotion modes, which is functionally added for simplicity and yet to optimize for fewer actuation units and better maneuver capabilities. 
According to the previous works, achieving all-terrain locomotion modes is still a challenging design gap and a direction for unraveling a triphibian robot design. Thus, we contributed to the triphibian robot design.} In previous work\cite{prev_work}, we have already presented a rolling hovercraft for amphibious locomotions which can be used both on land and on the water surface. 

\par Based on \textcolor{black}{our} previous amphibious robot, we \textcolor{black}{further} design a new triphibian platform, which opts for all-terrain operation with the same group of actuators. As mentioned in \cite{hybrid-model-sim}, the same propulsion system in different media (air and water for example) leads to inefficiency for at least one of these media. Therefore, in our design, to improve efficiency, we utilized \textcolor{black}{the propulsion system same to \cite{prev_work}} with a tri-rotor mechanism with ducted propellers operating in the air to provide the driving force on land, on the aquatic surface, and in the air. A cylindrical body rolls as a wheel on land, and a morphable mechanism assists the robot to transit from land/aquatic surface mode to aerial mode, and vice versa. 

The paper outline is summarized as follows: Section II introduces our platform design, Section III analyzes the different locomotion modes, and transitions between different modes are presented in Section IV. Experiment verification and validations are detailed in Section V and our contributions are summarized in the conclusion parts of Section VI.
\begin{figure}[hbp]
\centering
\includegraphics[{width=0.9\linewidth}]{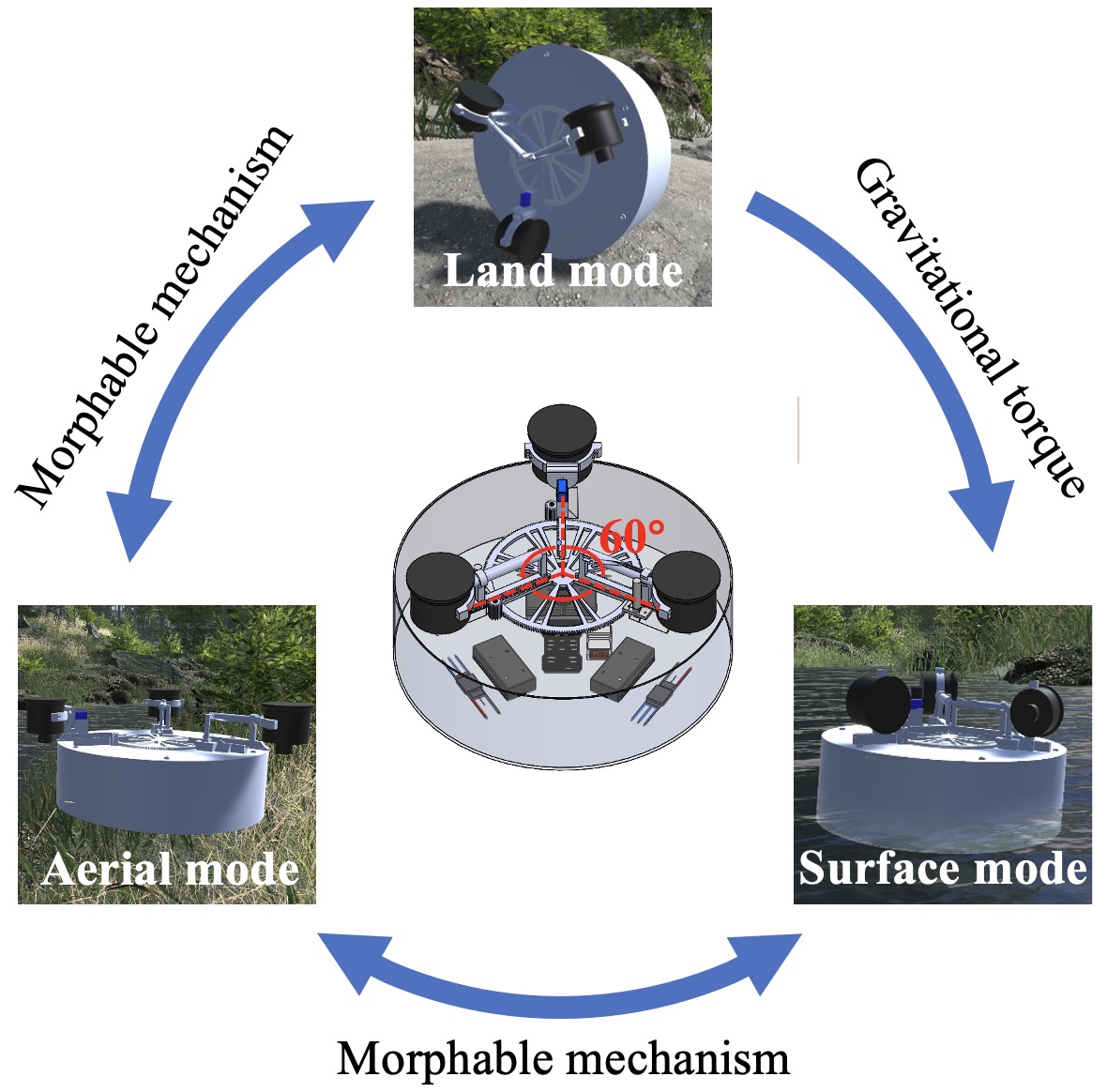}
\caption{Appearance of the vehicle with its propulsion and morphable state with its expansion mechanism.}
\label{fig-overview-robot}
\end{figure}

\section{Robot Design}
In this section, we first introduced the proposed robot's mechanical design, including the robot body and its actuation parts. Next, we detailed the robot \textcolor{black}{propulsion} system with its electrical parts together with the actuation unit selection to provide propulsion. Finally, \textcolor{black}{the morphable mechanism and its working principle were presented}.
\begin{table}[]
	\centering
	\caption{Robot Specifications}
	\label{tab-robot-spec}
	\resizebox{\linewidth}{!}{%
		\begin{tabular}{cc}
			\hline
			\textbf{Item}  & \textbf{Specifications}    \\ \hline
			Diameter       & 429 mm                      \\
			Height         & 130 mm (body height), 94 mm (deformation height)                     \\
			Weight         & 4.76 kg                      \\
			Controller     & Pixhawk autopilot                                  \\
			Operation mode     & Automatic or manual control                                 \\
			Modes          & Land, surface, aerial mode        \\
			Power          & \textcolor{black}{Lithium-ion Bat.} (6S 3500mAh 35C), 80A ESCs \\
			Actuation unit & ducted propellers, 70 mm 12 blades          \\ \hline
		\end{tabular}%
	}
\end{table}
\subsection{Robot Mechanism Design}
Our triphibian robot enables operations on the land, on the water surface, and in the air with \textcolor{black}{appropriate} mechanism design. As shown in Fig. \ref{fig-overview-robot} and Table \ref{tab-robot-spec}, the robot's mechanical parts mainly consist of a cylindrical body and a morphable mechanism together with its actuation units. For land mode, \textcolor{black}{the efficiency is improved} by adopting the cylindrical body as a rolling mechanism on the ground. For \textcolor{black}{surface mode, the cylindrical body} provides room to house and to secure all the electrical components including the flight controller, batteries, \textcolor{black}{ESCs,} transition servo, and the wireless receiver with waterproof requirements. For the tail servo and the ducted propellers, they are waterproof and exposed outside the hull. To achieve \textcolor{black}{land-surface mode} transition, these electrical components are packaged \textcolor{black}{ as shown in Fig. \ref{fig-overview-robot}} to ensure that the center of gravity is concentrated at one end away from the ducted propellers. The detailed reasons would be provided later. Besides, we further require synergized cooperation with the morphable mechanism, which is detailed in section C. 

\textcolor{black}{As for} the actuation units. As shown in Fig. \ref{fig-overview-robot}, we implement a tri-rotor configuration. Three motors are assembled equally at the robot boundary. This tri-rotor design is a benefit for the aerial mode since tri-rotor designs are more efficient in terms of compact size and power requirement \cite{tri-rotor-design}. 


\subsection{Propulsion System}
\begin{figure}[htp]
	\centering
	\includegraphics[{width=1\linewidth}]{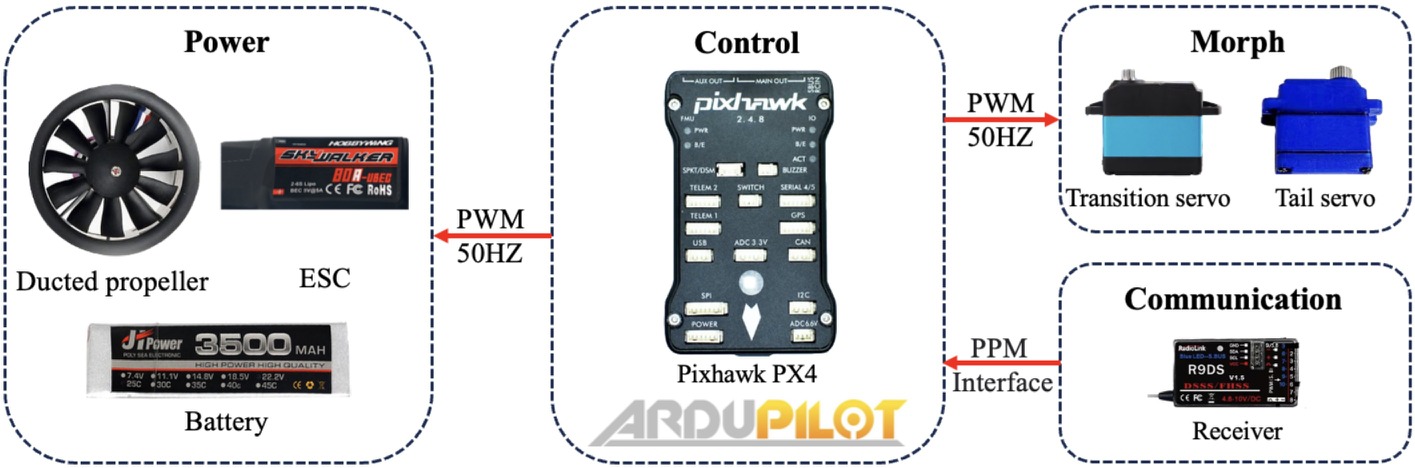}
	\caption{Robot electrical schematic diagram and its wiring overview.}
	\label{fig-electric}
\end{figure}

In this part, we detailed the robot propulsion system. First, as shown in Fig. \ref{fig-electric}, our system's electrical parts consist of three \textcolor{black}{Lithium-ion batteries} providing power for the whole system and deducted propellers. Electrical parts specifications can be found in Table \ref{tab-robot-spec}. A Pixhawk suite is employed as the main controller with the ArduPilot autopilot system \cite{ardupilot}. A receiver receives commands from the ground station for or wireless remoter. The tail servo is dynamically tilted to resist the imbalanced torque generated in the aerial mode for the tri-rotor setup design. The transition servo provides the driving force for the morphable mechanism to expand or retract its ducted propellers. Three ducted propellers provide propulsion for the robot. The weight of each ducted propeller is 178g, the maximum thrust of each is 2240g and their blade can only rotate \textcolor{black}{in one direction}. The ducted propellers are selected as the propulsion system actuation units for their higher hovering efficiency \cite{ducted-fans}. Compared with traditional propellers, ducted propellers have high aerodynamic efficiency. Therefore, the robot could be more compact with its weight and area further reduced with reasonable thrust requirement. In each mode, our allocation for each actuation ducted fan varied respectively. The details of allocation methods will be illustrated in Section \ref{sec-loco-mode}.
\par For the wiring connection in our proposed robot, three ducted propellers receive commands from three different Pixhawk PWM output channels. The tail servo and transition servo are connected to another two Pixhawk PWM output channels respectively. Manual and automatic control can be \textcolor{black}{achieved} by a wireless receiver linked to the Pixhawk. 
\begin{figure*}[htp]
	\centering
	\includegraphics[{width=\linewidth}]{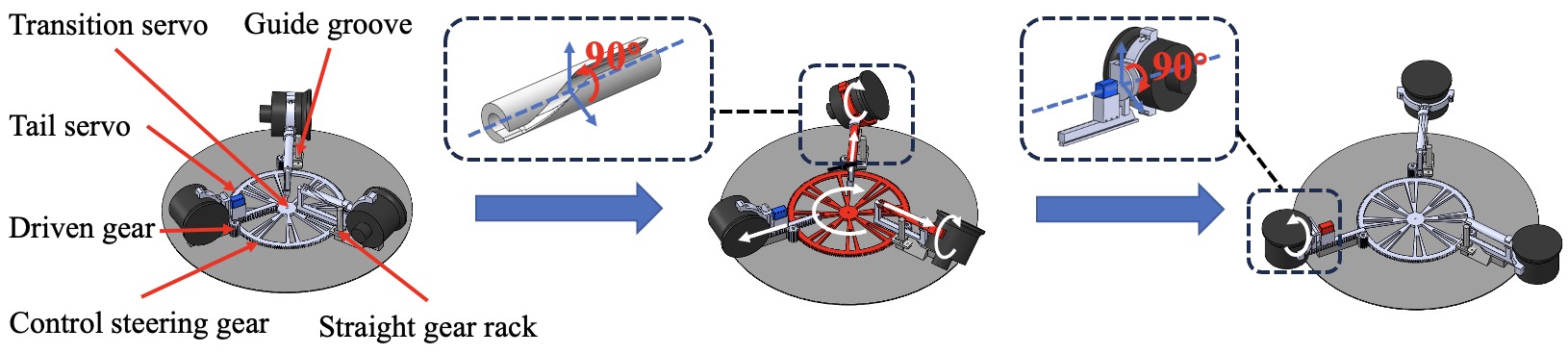}
	\caption{Morphable mechanism from surface mode (ducted propellers retraced within the robot body) to aerial mode (ducted propellers expanded outside the robot body).}
	\label{fig-morphable}
\end{figure*}
\subsection{Morphable Mechanism}
Our robot employs one set \textcolor{black}{of} propulsion system with a morphable mechanism for locomotion transitions. As shown in Fig. \ref{fig-morphable}, in land or \textcolor{black}{surface mode}s, the ducted propellers are \textcolor{black}{in retracted state}. For the aerial mode, the ducted propellers are \textcolor{black}{in extended state} to provide lift forces to fly. The morphable mechanism includes a transition servo connected with a control steering gear, three driven gears, three straight gear racks assembled in their slots, and two guide grooves. For the modes transition, the morphable mechanism achieves two goals: (1) pushing the ducted propellers outside the robot body. (2) adjusting the ducted propeller's thruster direction from the horizontal along the robot's top surface to the perpendicular direction of the top surface.
\par For the morphable mechanism working procedures, as shown in Fig. \ref{fig-morphable}, the transition servo controls the control steering wheel to rotate, and driven gears rotate along with the control steering wheel. Then the straight gear rack expands along the sliding slots and, the two grooves start to function. The cut-out guidance channel of the guide groove is composed of a straight part joined by a \textcolor{black}{90°} circular change and the other straight part (Fig. \ref{fig-morphable}).  Therefore, the ducted propellers are firstly expanded outside the robot body, followed by a rotation operation to achieve the thruster forces direction adjustment. \textcolor{black}{The two straight parts serve a dual purpose in expanding the ducted propellers while ensuring their stability.} The length of each part is designed according to the size of the hull, the gears and ducted propellers.
Besides, since ducted propellers are arranged in two different directions, to rotate all ducted propellers upwards, two opposite rotation directions are needed. 
\par The two ducted propellers rotating are realized by guide grooves (one rotating counterclockwise and one rotating clockwise) while the one rotating clockwise is controlled by the tail servo. 
The main reason why we use the tail servo control to realize ducted propeller rotation is that the tail servo used in aerial mode can free us from using another guide groove. \textcolor{black}{Since the tail servo has a rotation of 180°, the ducted propeller bonded to it can rotate 180° instead of 90°.} The use of the tail servo in the morphable mechanism avoids the increase in the total weight of the robot since no more parts such as guide grooves are needed for the bounded propeller. 
\begin{figure}
	\centering
	\includegraphics[{width=\linewidth}]{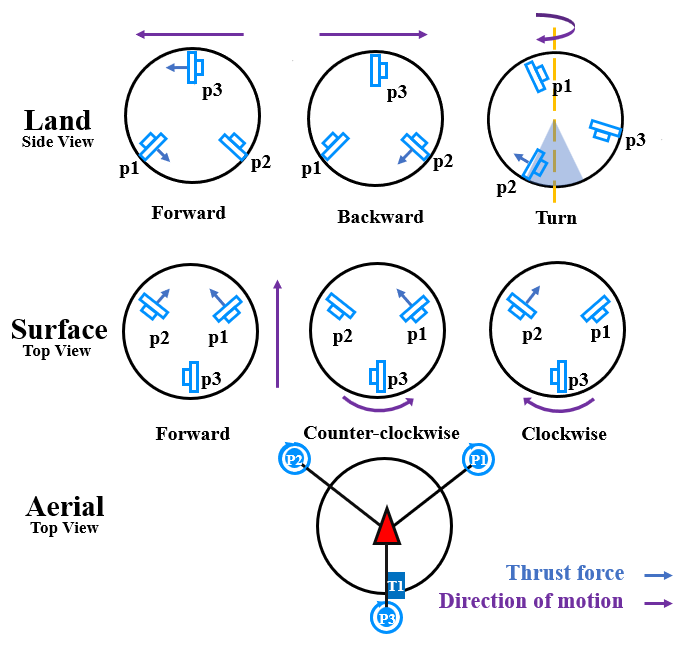}
	\caption{Different proposed motion modes: land mode, surface mode and aerial mode from top to bottom.}
	\label{fig-motion-all}
\end{figure}
\section{Modes of Locomotion}\label{sec-loco-mode}
\begin{figure}[h]
	\centering
	\includegraphics[{width=0.85\linewidth}]{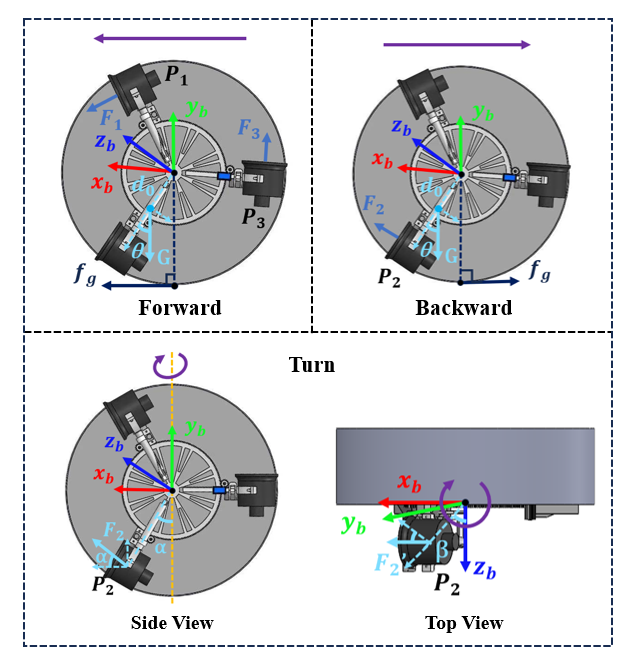}
	\caption{Land mode: rolling motion and turning motion forces and torques analysis.}
	\label{fig-force-land}
\end{figure}

The triphibian robot has three locomotion modes, land mode, aquatic surface mode, and aerial mode as shown in Fig. \ref{fig-motion-all}. In different locomotion modes, the robot adopts the propulsion system to cooperate with the proper morphable mechanism state. In this section, we specify the three modes with their respective actuation allocation and morphable mechanism separately. 
One propulsion system with a proper thruster allocation method for various locomotion modes could improve the utilization rate and reduce the weight of the whole system. 
To simplify our explanation, we set a coordinate system for the robot (Fig. \ref{fig-force-land}), which is located at the actuator centers formed plane $X_b$$Y_b$. The $X_b$ axis is defined by pointing from the center of the cover $O_b$ to the head of the robot. The ducted propeller P3 is on the negative $X_b$ axis. 
The $Z_b$ axis is perpendicular to the $X_b$$Y_b$ plane with the right-hand rule, pointing to the robot's outside part.

\begin{table}[]
	\centering
	\caption{Actuation units in various locomotion modes}
	\label{tab-loco-mode}
        \begin{threeparttable}
	\resizebox{\linewidth}{!}{%
		\begin{tabular}{cccccc}
			\hline
			\multirow{2}{*}{\begin{tabular}[c]{@{}c@{}}Locomotion\\ modes\end{tabular}} & \multirow{2}{*}{Operations} & \multicolumn{4}{c}{Actuation units} \\ \cline{3-6} 
			&                   & P1 & P2 & P3 & T1 \\ \hline
			\multirow{3}{*}{Land}    & Forward           & f1  & 0 & f3 & 0   \\
			& Backward          & 0 & f2  & 0  & 0  \\
			& Turn & 0 & f2  & 0  & 0  \\ \hline
			\multirow{3}{*}{Surface} & Forward           & f1 & f2  & 0 & 0  \\
			& Counter-clockwise         & f1  & 0  & 0 & 0  \\
			& Clockwise & 0 & f2  & 0  & 0  \\ \hline
			Aerial & Take-off          & f1 & f2 & f3 & 1  \\ \hline
		\end{tabular}%
	}
         \begin{tablenotes}
        \footnotesize
        \item P1, P2, and P3 are propellers. T1 is the tail servo. 
      \end{tablenotes}
  \end{threeparttable}

\end{table}


\subsection{Land mode}\label{AA}
For land mode, our robot can move forward, backward and turn. Similar to \cite{land-air-flystar}, the robot itself acts as a wheel by rolling motions. To ensure smooth rolling, the morphable mechanism retracts, and the three actuation ducted propellers inscribe within the robot body. As shown in Fig. \ref{fig-motion-all}, the main difference between forward motion, backward motion, and clockwise motion is caused by the allocation of the actuation units and the position ratio of the center of gravity of the robot relative to the ground. For each mode, the three actuators work together in different ways to generate the power of movement. Detailed information on the cooperation of three ducted propellers is shown in Table. \ref{tab-loco-mode}.
In forward motion (Fig. \ref{fig-force-land}-subplot Forward), the two active ducted propellers ($P_1$ and $P_3$) rotate in the same direction and the ducted propeller $P_2$ stays still. As shown in Fig. \ref{fig-force-land}, the two ducted propellers provide two forces $F_1$, $F_3$ to the vehicle to make the whole robot rotate along the $z_b$ axis. To simplify the model analysis, we only consider the forces on the $X_b$$Y_b$ plane and the gravity force $G$. We assume ground friction to be $f_g$. Thus, we have:
\begin{equation}
    F_1  d_1 + F_3  d_3 - f_g  d_g + G  d_0 \sin{\theta}= I_z  \omega_z \label{forward_eq}
\end{equation}
where $d_1$ is the distance between the ducted propeller $P_1$ and the center, $d_3$ is the distance between the ducted propeller $P_3$ and the center, $d_g$ is the distance between the ground friction and the center, $d_0$ is the distance between the center of gravity and the center, $I_z$ is the rotational inertia, $\omega_z$ is the angular acceleration of the vehicle. 

In backward motion (Fig. \ref{fig-force-land} Backward), the ducted propellers are in the opposite state compared with the forward mode. The two ducted propellers ($P_1$ and $P_3$) stay still and the ducted propeller $P_2$ works. Similar to the forward motion, the torque applied to the robot is along the $z_b$ axis. The only difference between these two torques is the direction. One makes the robot rotate counterclockwise while the other makes the robot rotate clockwise. Thus, the robot can achieve backward rolling. Here is the formula. The physical meanings of the elements are similar to those of forward motion:
\begin{equation}
    F_2  d_2 - f_g d_g - G  d_0  \sin{\theta}= I_z  \omega_z \label{backward_eq}
\end{equation}
It is worth noticing that by switching between forward motion and back motion, the robot could stop rolling with proper control.   

In turn motion, the vehicle would rotate in the direction perpendicular to the $X_b$$Y_b$ plane. As shown in Fig. \ref{fig-force-land} Side View and Top View, the ducted propeller $P_2$ is close to the center of gravity of the vehicle so that the ducted propeller $P_2$ could remain in the lower part when turning from rest and the robot could turn with less power. The component of $F_2$, which is perpendicular to the $X_b$$Y_b$ plane, would provide torque for the robot to turn.
If we project the force of diagram Side View onto the angle of diagram Top View, we get $F_2^\prime$ equals to $F_2$ times $\cos{\alpha}$. From the perspective of Figure Top View, we get the following equation: 

\begin{gather}
	F_2^\prime = F_2 \cos{\alpha}\\
        d^\prime = \frac{d}{\cos{\beta}} \\
	F_2^\prime \cos{\beta} d^\prime   = I \omega \label{turn_eq}
\end{gather}
Thus, the simplified equation become:

\begin{equation}
    F_2 d\cos{\alpha}    = I  \omega \label{turn_eq}
\end{equation}
where $d$ is the vertical distance between the force $F_2$ and the center, $I$ is the rotational inertia, $a$ is the angular acceleration of the vehicle.
From the above equation, the smaller the angle $\alpha$, the larger the $\cos{\alpha}$, the easier for the robot to turn. 
Meanwhile, for the rolling in forward direction, the robot also follows Equation (\ref{forward_eq}). However, the distance between the ducted propeller $P_2$ and the center of gravity is close, the moment along the $Z_b$ axis is not large enough. The ground friction along this direction would also cancel the rolling torque provided by the $F_2$.

One thing to be specified is the difference between the backward motion and the turn motion. Although both motions use a single ducted propeller to drive the robot, the starting positions of the two modes are different. For backward motion, it can be realized in some parts where the center of gravity is away from the ground as shown in Fig. \ref{fig-motion-all}. For the turn motion, it can only be achieved by providing an accurate abrupt force (should not be too large that the robot could roll on the $X_b$$Y_b$ plane) when the ducted propeller $P_2$ is close to the ground. The reason could be found in Equation (\ref{turn_eq}). The smaller the angle $\alpha$, larger the value of $\cos\alpha$, the easier to turn. 


\begin{figure}[h]
	\centering
	\includegraphics[{width=1\linewidth}]{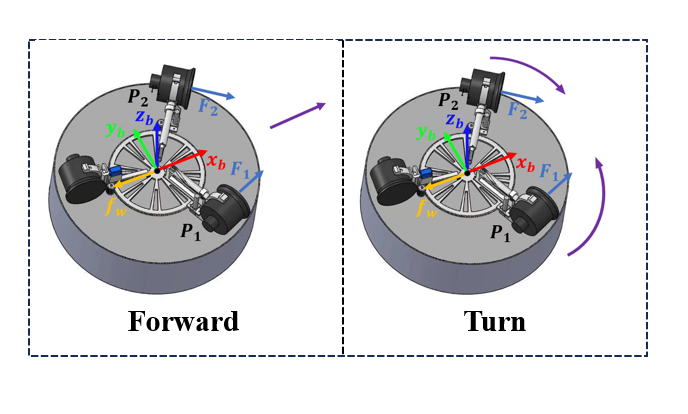}
	\caption{Surface mode: forward and turning motion forces and torques analysis.}
	\label{force-surface}
\end{figure}
\subsection{Surface mode}
As shown in Fig. \ref{force-surface}, the three ducted propellers are above the water in surface mode. The torques generated by $P_1$ or $P_3$ can cause the robot to rotate counterclockwise, and the torque provided by $P_2$ can cause the robot to rotate clockwise. Thus, we use different ducted propellers to control the movement of the robot in different modes. The details of actuation units can be seen in Table. \ref{tab-loco-mode}. 

In forward motion, the two ducted propellers $P_1$ and $P_2$ are activated while ducted propellers $P_3$ is off. As shown in Fig. \ref{force-surface}, ducted propeller $P_1$ provides $F_1$ and $P_2$ provides $F_2$. The magnitudes of $F_1$ and $F_2$ are the same and these two forces make the same angle (\textcolor{black}{60°}) with the $X_b$ axis. In this case, the two ducted propellers can work together to generate a driving force in the surge direction and cancel the force components in the sway direction to avoid rotation. Here we can calculate the acceleration of the robot:
\begin{equation}
    F_{1}  \cos{60^\circ} + F_{2}  \cos{60^\circ} - f_w  = m a_{com}
\end{equation}
where $f_w$ is the resistant force, $a_{com}$ is the acceleration of the robot and $m$ is the weight of the robot.

In clockwise and anticlockwise turn motions, only one ducted propeller operates each time. Theoretically, the three propellers can all be used to generate torques so that the robot can rotate on the water's surface. As shown in Table. \ref{tab-loco-mode}, in our design, $P_2$ is used for clockwise rotation while $P_1$ is used for counterclockwise rotation. The rotational inertia is $I_{com}$. The combined external force generates a torque $\tau_e$. The angular acceleration of the vehicle $\omega_{com}$ in the pure rotation stage could be calculated through:
\begin{equation}
    F_{1}  d_1 + \tau_{e}  = I_{com} \omega_{com}
\end{equation}
or
\begin{equation}
    F_{2}  d_2 + \tau_{e}  = I_{com} \omega_{com}
\end{equation}
\begin{figure}[htp]
	\centering
	\includegraphics[{width=0.55\linewidth}]{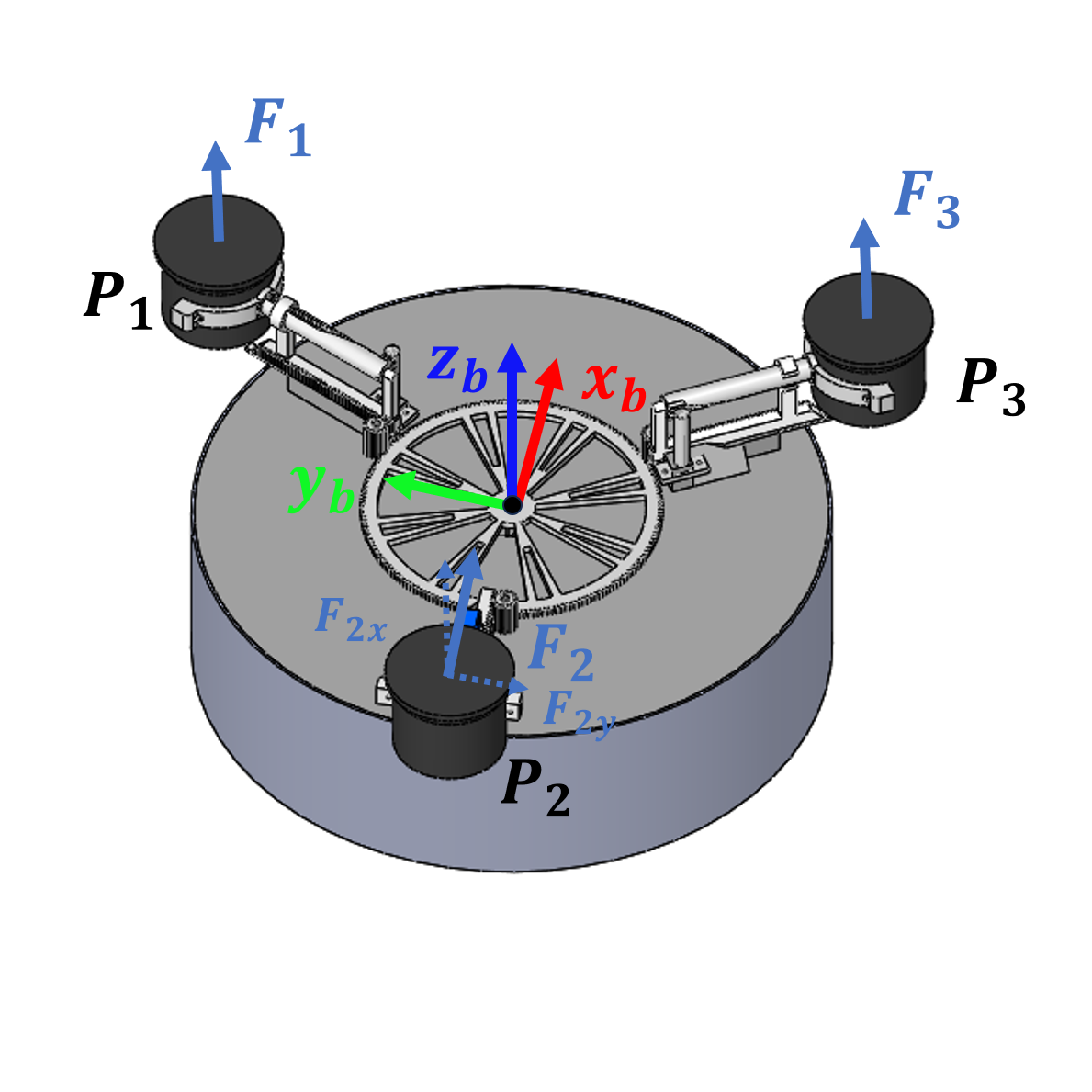}
	\caption{Aerial mode: lifting forces and torques analysis with its extended actuation units.}
	\label{fig-force-aerial}
\end{figure}
\subsection{Aerial mode}


In aerial mode, our robot has a tri-rotor design. Compared with other multi-rotor Unmanned Aerial Vehicles (UAVs), tri-rotor UAVs are smaller in size, lighter in weight, and have no size limitation\cite{adv_of_tri}.
Thus, for aerial mode, we used the tri-rotor model of the ArduPilot \textcolor{black}{firmware} \cite{ardupilot} to control our robot. 

As shown in Fig. \ref{fig-force-aerial}, the three ducted propellers' blades all rotate counterclockwise. The ducted propellers $P_1$ and $P_3$ are vertically fixed on the hull and ducted propellers $P_2$ can tilt left and right around with the help of the tail servo. The component force $F_{2y}$ that $P_2$ provided along the $y_b$ axis generates a torque. When hovering in the air, the magnitude of this torque is equal to the anti-torque generated by the three ducted propellers and the directions of these two torques are opposite. \cite{adv_of_tri} When the robot turns left or turns right in the air, the torque helps the robot to adjust its direction.

\section{Transition between modes}
For transition between modes, we first need to deal with is the waterproofing of electronic devices. When the robot rolls to the water surface, the gap between the robot's cover and the hull would be in contact with the water. When taking off from the water, the wind generated by the ducted propellers will roll up a great amount of water. Thus, waterproofing is necessary for the robot. We use two-layer structure for the cover and add some rubber gasket in the gap so that the whole robot is water-tight.

To achieve the all-terrain locomotion in different media, we proposed the transition methods between different modes as shown in Fig. \ref{exp-diff-modes}. We detailed the different roles of the center of gravity units and the morphable mechanism in this part.



\subsection{Land-surface transition}

\begin{figure}[h]
	\centering
	\includegraphics[{width=0.7
		\linewidth}]{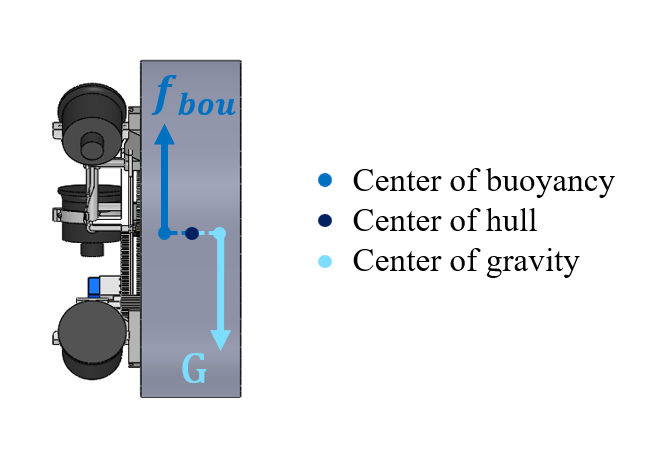}
	\caption{main view of the center of gravity on the ground.}
	\label{fig10}
\end{figure}

For the land mode, the robot roll as a wheel. However, the robot serves as a hovercraft in surface mode, and all the ducted propellers are assembled away from the water. In this case, we need to ensure the robot has flipped upward during the land-surface transition. In our design, we achieve this through the imbalance of the center of gravity and the centroid of the whole body. As shown in Fig. \ref{fig10}, the center of gravity is closer to the other side of the actuation unit. The gravity provides torque for the ducted propellers to flip upwards. Moreover, we stick some EPE (Expanded Polyethylene) particles, which have low density and are resistant to water, on the side where the ducted propellers are. In this case, the weight of the water displaced by the object would be larger for the half where the actuators lie than the other half of the hull. The buoyancy force can be seen as acting at the first half of the hull like Fig. \ref{fig10} and it provides torque to flip the ducted propellers upwards. Therefore, when the robot is switching from land mode to surface mode, the actuation unit tends to flip upwards and the side where the center of gravity is closer tends to flip downwards. Thus, we can successfully finish the switch between land mode and surface mode. 
\subsection{Surface-aerial transition}

For the surface mode, the forces provided by the three ducted propellers are all parallel to the plane formed by $X_b$ and $Y_b$ (Fig. \ref{force-surface}). However, for aerial mode, these forces should be perpendicular to the $X_b$$Y_b$ plane. This is the first difficulty we need to solve. Here we use the morphable mechanism design to help. In our design, as we illustrate in Section II, the transition servo provides the force for almost the whole morphable mechanism, including the guide grooves that cause the ducted propeller $P_1$ and $P_3$ to rotate and extend. The tail servo guides the ducted propeller $P_2$'s transition alone. 
The second difficulty comes from the fact that fluid density in air and water have great variation. 
The density of water far exceeding that of air will greatly reduce the thrust of the ducted propellers, so that the thrust cannot make the robot leave the water.
In our design, we sidestep this problem with the design of the hovercraft in surface mode. The robot no longer needs to deal with such density variance when transiting between surface mode and land mode.


\section{Experiments and Analysis}
\begin{figure}[!htbp]
	\centering
	\includegraphics[{width=\linewidth}]{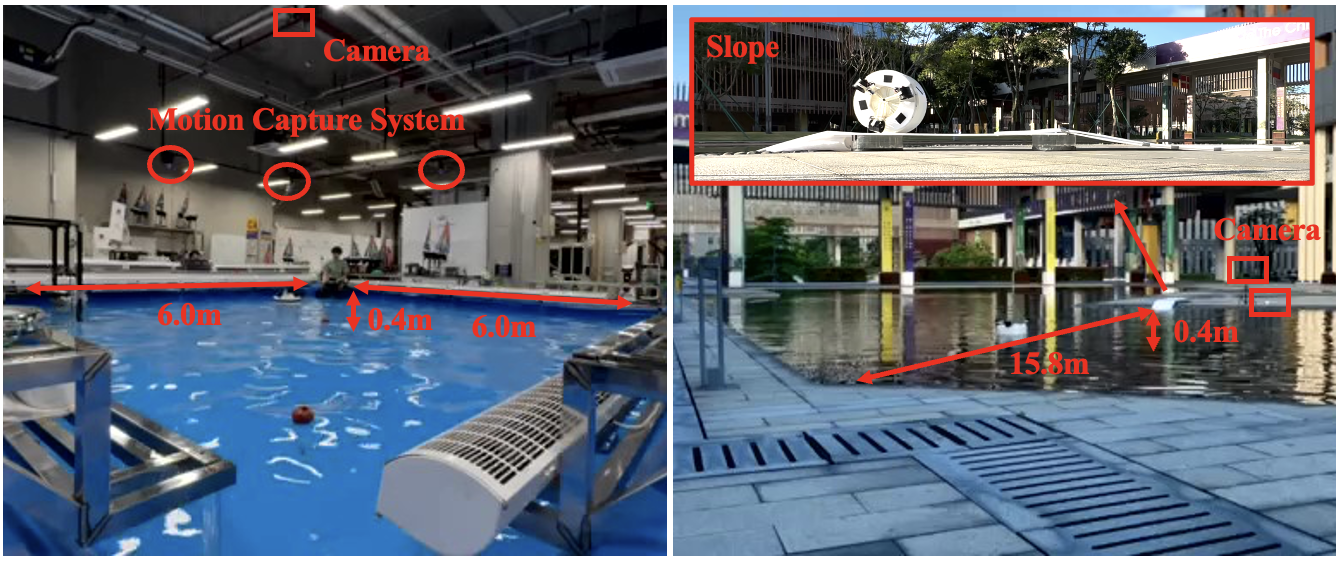}
	\caption{Experiment scenarios setup: Left: a 6m $\times$ 6m indoor pool; Right: an outdoor lake at The Chinese University of Hong Kong, Shenzhen, Guangdong, China.}
	\label{fig-exp-scenarios}
\end{figure}

In this section, we present the results of the experiments conducted on the robot. First, we tested the robot with its land mode, surface mode, and aerial mode separately. Next, we evaluated the robot's performance in a scenario where all modes are joined together. The experiments were conducted in a 6m $\times$ 6m $\times$ 0.4m indoor pool with the OptiTrack Motion Capture system and cameras around or an outdoor lake with cameras around and a white slope in Shenzhen, China (Fig. \ref{fig-exp-scenarios}). 


\begin{figure}[!h]
\centering
\includegraphics[{width=\linewidth}]{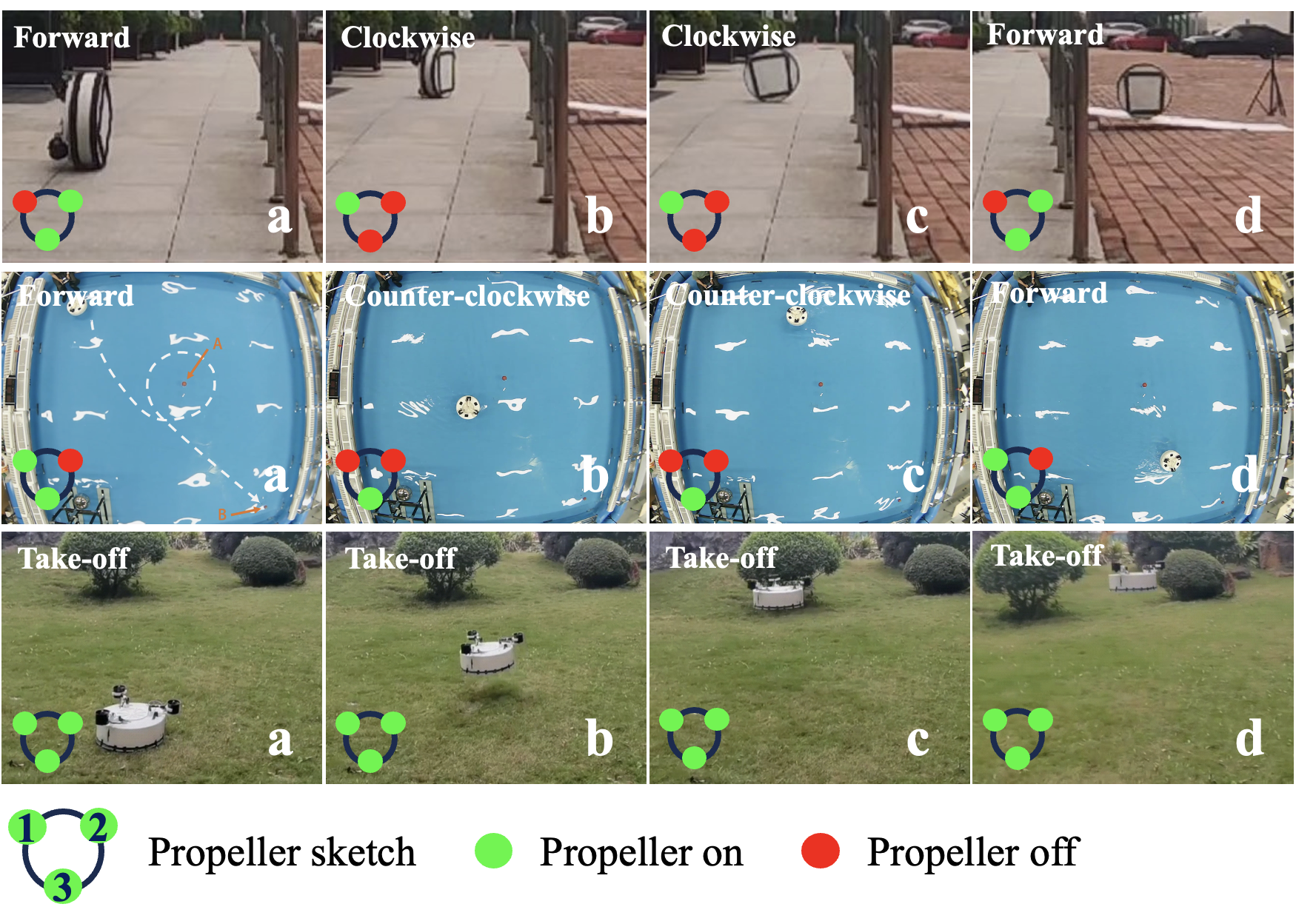}
\caption{Verification experiments snapshots for different motion modes: land mode, surface mode, aerial mode (from top to bottom row).}
\label{exp-diff-modes}
\end{figure}


\subsection{Analysis of each mode}
As shown in Fig. \ref{exp-diff-modes}, our robot can achieve good capabilities in three different motion modes: land modes (top row), surface mode (middle row) and aerial mode (bottom row). In all three different motion modes, the actuation units vary accordingly. In order to illustrate its working principle, we used a propeller sketch to indicate its on/off status for each propeller in different working modes. Green stands for "on" and red for "off", labelled in the bottom left corner during the working mode demonstration.  

\subsubsection{Land mode}
In land mode, the robot is supposed to operate on the land performing moving capabilities as a mobile robot, such as forwarding, backwarding, and turning around (clockwise or counter-clockwise). To showcase its land capabilities, the robot is supposed to follow a route and turn its direction to finish the land mode demonstration.
\par As shown in Fig. \ref{exp-diff-modes} top row, (a) part: the robot went forwarding climbing along a slope with \textcolor{black}{two propellers P1 and P3 on and one P2 off generating a driving force causing the robot to rotate along $Z_b$ axis and move along the routine, as analyzed in Fig. 6. For (b) and (c) part, the robot reached the turning position and performed the turning motion with P2 propeller on and P1, P3 off. This propeller generated forces upon on the ground generated a resultant force and thus a torque to cause the robot turn clockwise, as analyzed in Fig. 6. Finally, the robot continues to move on and rolls down a slope in (d) part to perform forwarding motion with similar propellers allocation to (a) case. Besides, the robot could move freely on rougher ground and smoother slopes, and more experiments on various ground surfaces (concrete floor, sports track, and bricks) are demonstrated in the supplementary experiments video, showing the robot's good moving capabilities. }


\subsubsection{Surface mode}
In this part, the robot surface mode is validated by performing its mobility on water. We tested the robot indoors with a task as shown in Fig. \ref{exp-diff-modes}: one obstacle in the center labelled as A in red, the other target at the corner labelled as B in red. In the experiment, the robot was supposed to move forward to circle the obstacle A at the center of the pool, and then continued to move forward to the target B destination position. The whole robot experiment movement path (recorded by Motion Capture system and cameras) has been recorded and shown in the (a) figure in the middle row Fig. \ref{exp-diff-modes}. 
\par In the whole task, the robot first moved forward to the obstacle with propellers P1, P2 on and P3 off as shown in part (a). As explained in Section III, Fig. 7, the robot is laying flat and the two propellers resultant force drove the robot to move along $X_b$ direction, causing the robot to move forward. In (b) and (c), the robot performed counter-clockwise circling motion around the obstacle A, with propeller P1 on. After that, the robot continued to move forward with similar propellers configuration as (a) part, and finally reached the destination spot B. This task demonstrated the robot surface mobility, and more potential tasks could be validated if we used the proposed as a surface platform together with more adds-on sensors (cameras and so on).
\subsubsection{Aerial mode}
This part demonstrated the robot aerial mobility. The robot started from the ground and then lifted itself to the air achieving the taking off operation. We showcased its aerial mode with the bottom row in Fig. \ref{exp-diff-modes}.
\par As the propeller sketch shows, all the propellers are on during the take off operation from the snapshots (a-d). One thing to mention is that, all the three propeller could generate enough lifting forces for the robot to go up, but it will also cause the robot to rotate along the $X_b$ axis for the rotor rotating effects. Therefore, to compensate the resulted rotating motion, we further added one servo motor T1 as shown in Fig. 5 to reduce the improper rotation motion. And this newly-added servo mechanism is mainly learned from that strategy of an tri-rotor aircraft. The working principle is that Aerial mode as explained in Section III.C. To sum up the robot aerial mode, as shown in bottom row in Fig. \ref{exp-diff-modes}, the robot took off from the ground, and took off with the three propellers and the tail servo on at the same time. For the forwarding operation, we could achieve this by adjusting the propellers forces accordingly. Besides, by lowering the propellers rotating forces, it could land on the ground properly.

\subsection{Verification in a joined usage scenario}

\begin{figure}[htbp]
\centering
\includegraphics[{width=\linewidth}]{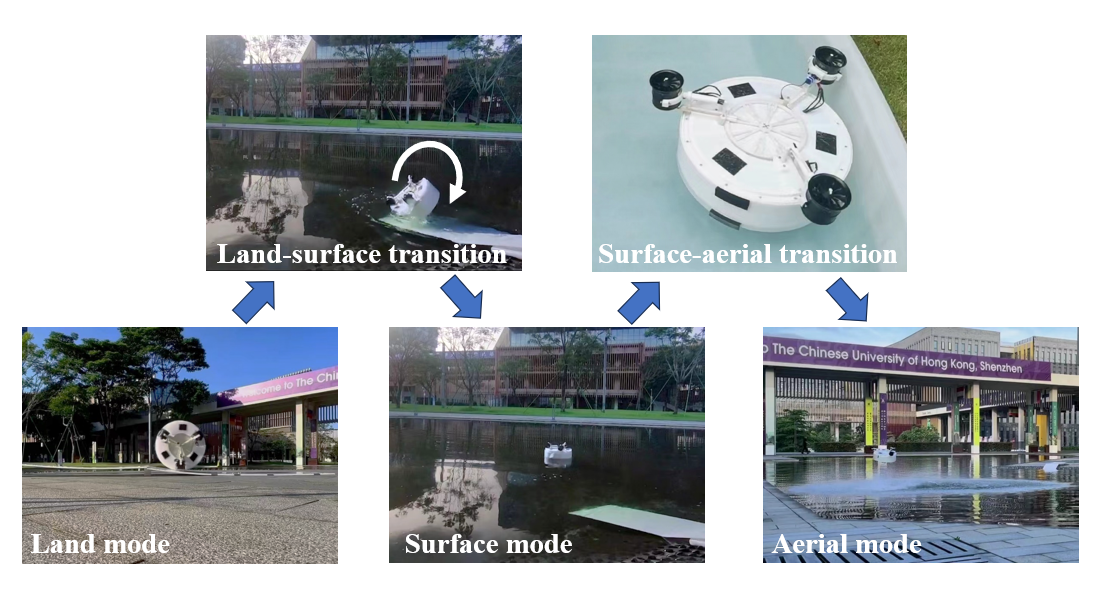}
\caption{Verification experiment of the transition example: from land mode to aerial mode, including land-surface transition and surface-aerial transition.}
\label{exp-trans}
\end{figure}

All the three working modes have been analyzed as above-mentioned. This part explained the transition process of the three working modes. Two design factors contribute to the mode conversions as shown in Fig. 2: morphable mechanism and gravitational torque. We detailed the principles behind the two conversions as illustrated in Section II.C and Section IV. To showcase the robot triphilian capabilities, we demonstrated the robot in a scenario where the robot performed land mode, surface mode and aerial mode by running in the ground, moving in the surface, and flying in the sky. The transition procedures are highlighted in Fig. \ref{exp-trans} to be composed as the whole pipeline.
\par The whole transition experiments composed land mode, surface mode, aerial mode (addressed in the above parts) and two transition procedures: land-surface transition and surface-aerial transition, which will be addressed in detail in this part. The robot started with land mode on ground and went forward to a slope near the pool. Rolling down a slope, it entered the pool achieving the transition procedures as shown in Fig. 1 and 12. The land-surface transition is mainly achieved by the imbalanced robot mass distribution, and thus when the robot flipped over, it would flip to the heavier part causing the propellers to be on the robot top side as illustrated in Section IV.A. Consequently, the body is floating on the water with the propellers operating on surface. Then it performed surface motions with surface mode. When the robot wanted to take off. It performed the surface-aerial transition with the efficacy of the morphable mechanism as illustrated in Fig. 4 and Section II.c. As the transition servo worked, the control steering gear would work with the driven gear to move the straight gear rack along the guide groove to the outer side of the robot body and rotate the ducted propellers \textcolor{black}{90°} meanwhile as shown in Fig. 4. Finally, the robot switched it running mode from surface mode to aerial mode, and took off from the water pool and successfully landed on the ground to its destination.

  

\section{Conclusions}
In this paper, we proposed a novel triphibian robot, which can operate on the water surface, move on the ground, and fly in the air with the same actuation system and a novel morphable mechanism design. The multimodal transition capabilities owe to the novel mechanical designs, and with the triphibian mobility, the robot is potential for various tasks in the future. One thing to note is that, the transition from aerial mode to land mode still needs a proper strategy when the robot lands on the ground and performs rolling movements by erecting its body. This feature is within our future work, and more experiments will follow to further improve the robot capabilities as well as its performances. Furthermore, to make it a more general platform for future applications, automatic control is needed to replace the manually control for now. We will also design a mechanical device that allows amphibious robots to perform \textcolor{black}{90°} flips on the ground, thereby enabling the switching of three different motion modes in a closed loop and enhancing the exploration capability of the robots. This will make our robot more fully functional.


\end{document}